\documentclass[10pt,twocolumn,letterpaper]{article}

\usepackage{cvpr}
\usepackage{times}
\usepackage{epsfig}
\usepackage{graphicx}
\usepackage{amsmath}
\usepackage{amssymb}
\usepackage{enumitem}
\usepackage{multirow}

\usepackage[pagebackref=true,breaklinks=true,letterpaper=true,colorlinks,bookmarks=false]{hyperref}

\cvprfinalcopy 

\def\cvprPaperID{711} 

\ifcvprfinal\fi
\begin{document}

\makeatletter
\def\blfootnote{\gdef\@thefnmark{}\@footnotetext}
\makeatother
	
\title{SurfConv: Bridging 3D and 2D Convolution for RGBD Images}

\author{Hang Chu$^{1,2}$~~~Wei-Chiu Ma$^{3}$~~~Kaustav Kundu$^{1,2}$~~~Raquel Urtasun$^{1,2,3}$~~~Sanja Fidler$^{1,2}$\\
	$^1$University of Toronto~~~$^2$Vector Institute~~~$^3$Uber ATG\\
	{\tt\small \{chuhang1122, kkundu, fidler\}@cs.toronto.edu~~~~\{weichiu, urtasun\}@uber.com}
}

\maketitle

\begin{abstract}
	We tackle the problem of using 3D information in convolutional neural networks for down-stream recognition tasks.
	Using depth as an additional channel alongside the RGB input has the scale variance problem present in image convolution based approaches.
	On the other hand, 3D convolution wastes a large amount of memory on mostly unoccupied 3D space, which consists of only the surface visible to the sensor.
	Instead, we propose SurfConv, which ``slides'' compact 2D filters along the visible 3D surface.
	SurfConv is formulated as a simple depth-aware multi-scale 2D convolution, through a new Data-Driven Depth Discretization ($ D^4 $) scheme.
	We demonstrate the effectiveness of our method on indoor and outdoor 3D semantic segmentation datasets. Our method achieves state-of-the-art performance with less than 30\% parameters used by the 3D convolution-based approaches. \blfootnote{Code \& data at \url{https://github.com/chuhang/SurfConv}}
\end{abstract}

\section{Introduction}
While 3D sensors have been popular in the robotics community, they have gained prominence in the computer vision community in the recent years. This has been the effect of extensive interest in applications such as autonomous driving~\cite{geiger2013vision}, augmented reality~\cite{orts2016holoportation} and urban planning~\cite{wang2016torontocity}. These 3D sensors come in various forms such as active LIDAR sensors, structured light sensors, stereo cameras, time-of-flight cameras, \etc. These range sensors produce a 2D depth image, where the value at every pixel location corresponds to the distance traveled by a ray from the sensor through the pixel location, before it hits a visible surface in the 3D scene.



\begin{figure}[t!]
\begin{center}
\begin{tabular}{c}
\includegraphics[width=0.94\linewidth]{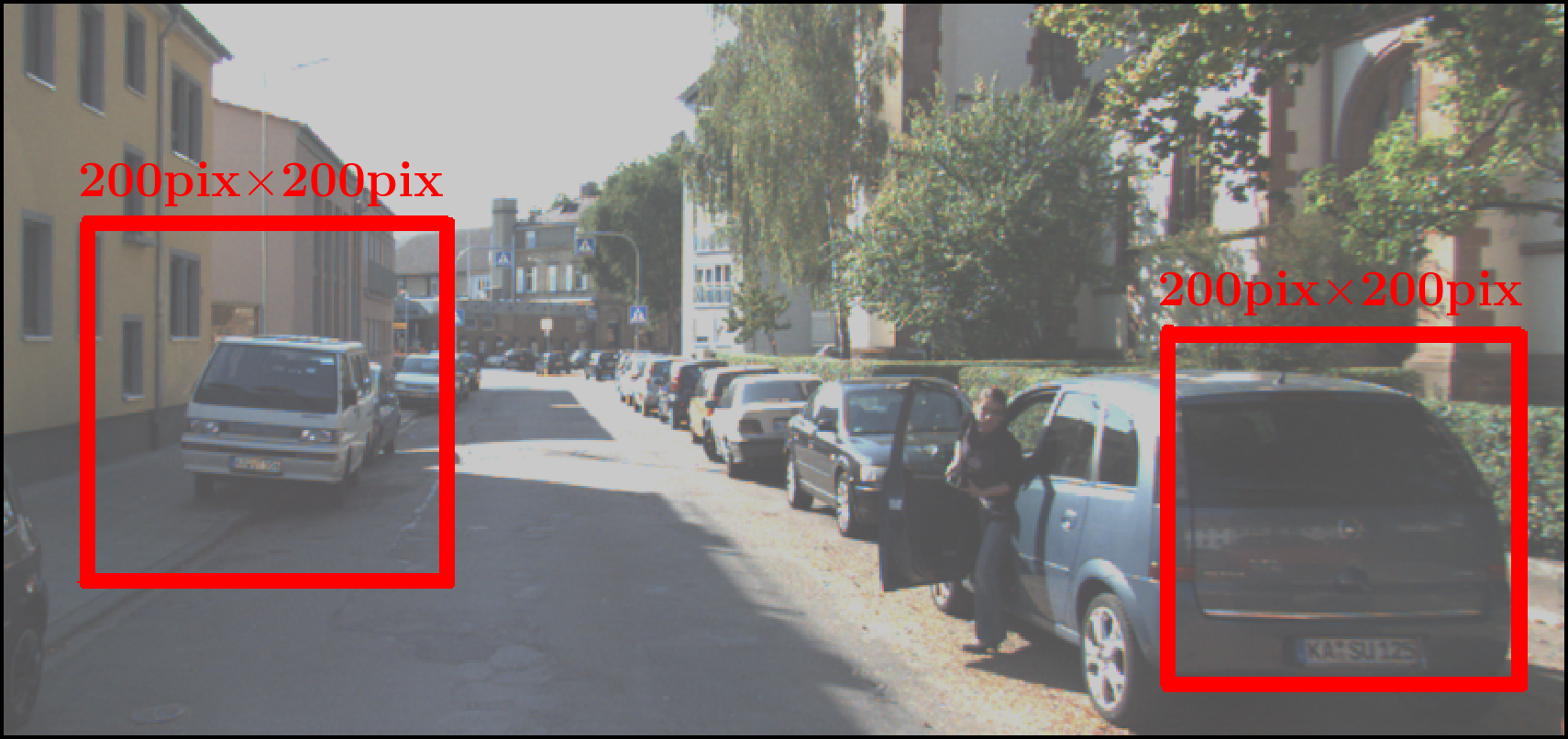}\\
Image Convolution\\
\includegraphics[width=0.94\linewidth]{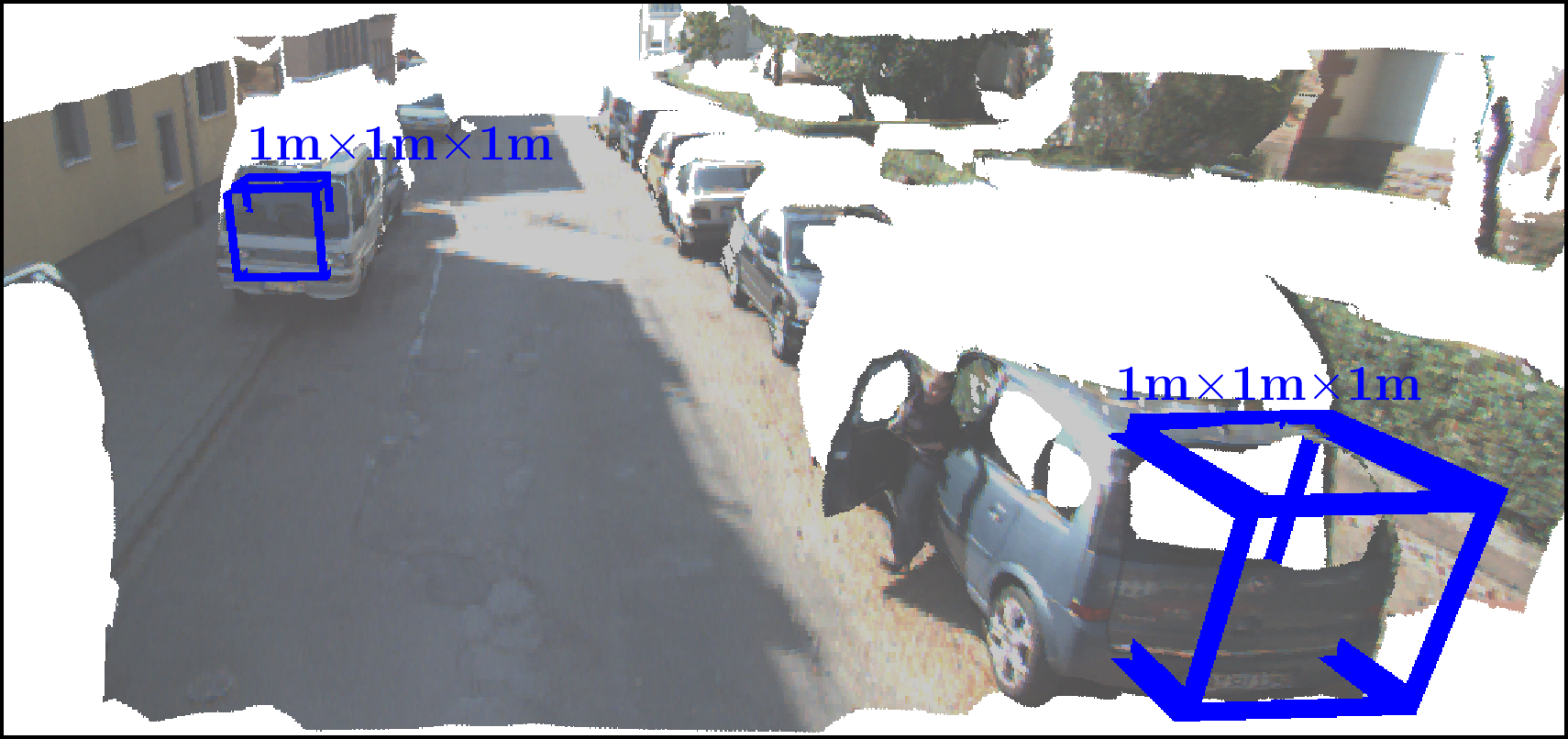}\\
3D Convolution\\
\includegraphics[width=0.94\linewidth]{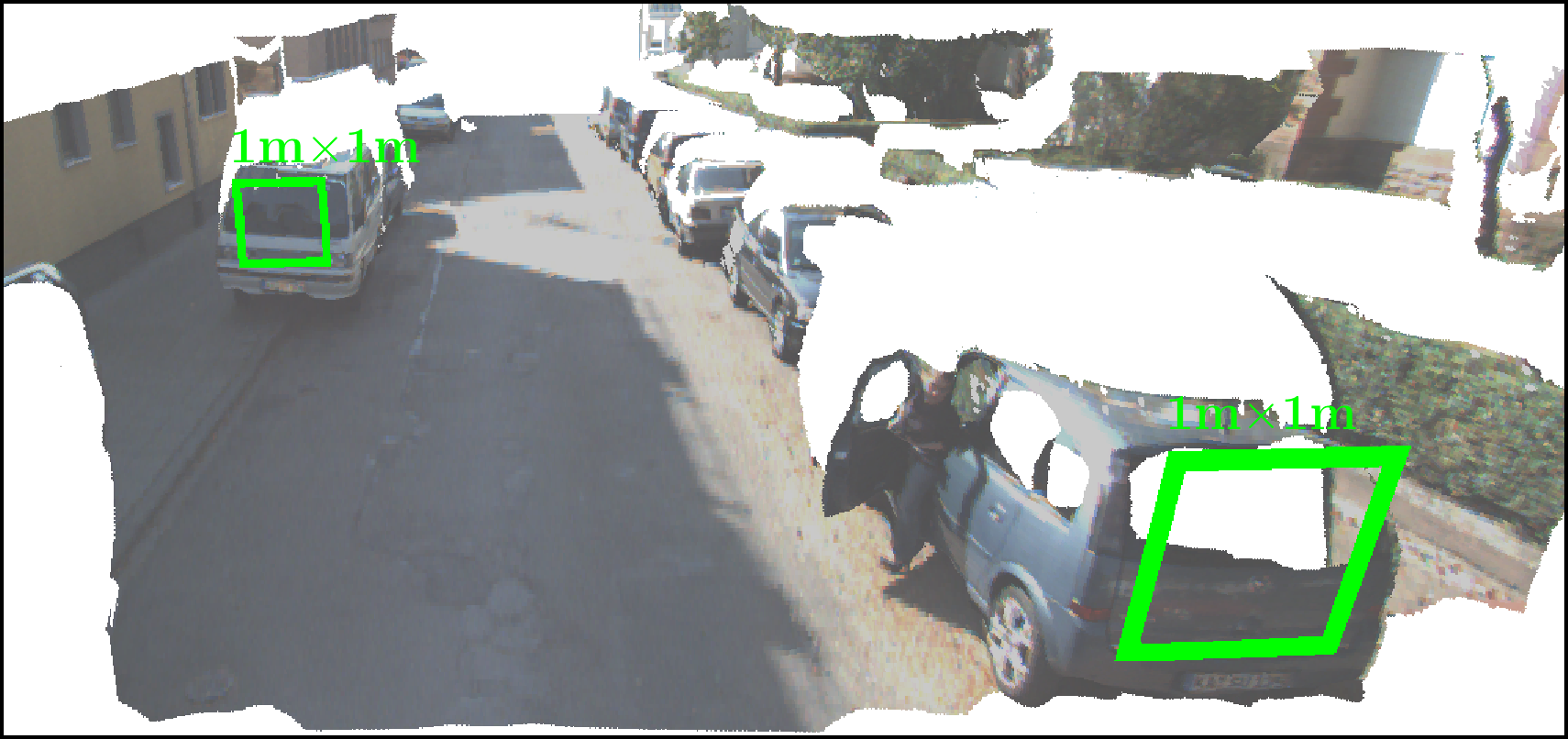}\\
Surface Convolution\\
\end{tabular}
\end{center}
\caption{\footnotesize A 3D sensor captures a surface at a single time frame. 2D image convolution does not utilize 3D information and suffers from scale variance. 3D convolution solves scale variance, but suffers from non-volumetric surface input where majority of voxels are empty. We propose surface convolution, that convolutes 2D filters along the 3D surface.}
\label{fig:intro}
\vspace{-0.4cm}
\end{figure}

Recent success of convolutional neural networks for RGB input images~\cite{krizhevsky2012imagenet} have raised interests in using them for depth data. One of the common approaches is to use handcrafted representations of the depth data and treat them as additional channels alongside the RGB input~\cite{gupta2014learning, fang20153d}. While this line of work has shown that additional depth input can improve performance on several tasks, it is not able to solve the scale variance problem of 2D convolutions. In the top of Fig.~\ref{fig:intro}, we can see that for two cars at different distances, the receptive fields of a point have the same size.
This means that models are required to learn to recognize the same object in different inputs.

To overcome this issue, an alternative is to represent the data as a 3D grid and use 3D convolution on it~\cite{wu20153d}. For such a dense representation, it requires huge computation and memory resources. This limits the resolution in all three dimensions. Furthermore, since 3D sensor captures the information of how far the objects are from the sensor at a single frame, the visible surface of the scene occludes the rest of the 3D volume. Thus, the information in the input occupies an extremely small fraction ($ \sim $ 0.35\%\footnote{Calculated with the standard 0.1m resolution for~\cite{geiger2013vision} and 0.02m resolution for~\cite{silberman2012indoor}}) of the entire volume.  This results in the 3D convolution based approaches to spend a large fraction of time and memory on the unoccupied empty space shown in the middle of Fig.~\ref{fig:intro}.

We propose to reformulate the default 3D convolution as \textit{Surface Convolution} (SurfConv) for a single frame RGBD input. Instead of ``sliding'' 3D filters in the voxel space, we slide compact 2D filters along the observed 3D surface. This helps us to exploit the surface nature of our input and help the network learn scale-invariant representations (bottom of Fig.~\ref{fig:intro}). A straight-forward implementation of surface convolutions is challenging since it requires depth-dependent rescaling at every location, which is a computational bottleneck. To address this problem, we propose a \textit{Data-Driven Depth Discretization} ($D^4$) scheme, which makes surface convolution practically feasible. We use our approach to show state-of-the-art results on the single-view 3D semantic segmentation task in KITTI~\cite{geiger2013vision} and NYUv2~\cite{silberman2012indoor} benchmarks. To summarize, our main contributions are:
\begin{itemize}
	\item We propose Surface Convolution, that processes single frame 3D data in accordance with its surface nature.
	\item We propose to realize Surface Convolution through a Data-Driven Depth Discretization scheme, which offers a simple yet effective solution that achieves state-of-the-art single view 3D semantic segmentation.
\end{itemize}

\section{Related Work}

\paragraph{Deep 2D RGBD Representations.} In the last few years, 2D CNNs have been used to create powerful feature descriptors for images~\cite{krizhevsky2012imagenet}, and can learn complex patterns in the data~\cite{zhou2014object}.
One of the approaches to extend the success of these 2D convolutions to range data, is by projecting the 3D data into multiple viewpoints, each of which is treated as a 2D input~\cite{chen2016multi, qi2016volumetric, shi2015deeppano, su2015multi}. However, the computation time scales linearly with the number of views. Since a single frame RGBD image sees only the unoccluded portion of the 3D world, the visible surfaces from drastically different viewpoints might not align with that of the input camera viewpoint. Furthermore, reasoning about multiple viewpoints does not lead to a natural, interpretable 3D description of the scene into parts and their spatial relations~\cite{hoiem2011representations}. 

Another alternative is to simply use handcrafted depth representations and treat them as additional channels alongside the RGB input. Such approaches have shown to improve for tasks such as 3D shape retrieval~\cite{fang20153d, xie2015deepshape}, semantic segmentation~\cite{guo20153d, gupta2014learning, li2016lstm, long2015fully} and 3D object detection~\cite{chen20173d, deng2017amodal, engelcke2017vote3deep, li2016vehicle}.

\paragraph{3D Convolution.} To handle the scale variance, 3D convolution learns the correlations directly in the 3D space.
To represent the point cloud information, input representations such as occupancy grids~\cite{huang2016point, maturana20153d, sedaghat2016orientation} and TSDF~\cite{dai2017scannet, ge20173d, song2016deep, zeng20163dmatch, zhang2016deepcontext} have been explored.

One of the key challenges in 3D convolution is the fact that increasing the input dimensions can lead to significant increase in the memory requirements. Thus, common practices are to either limit the input resolution to a low resolution grid, or reduce the network parameters~\cite{song2016semantic}. Since the range data is sparse in nature, approaches such as~\cite{engelcke2017vote3deep, riegler2016octnet,  uhrig2017sparsity} have also aimed at reducing the memory consumption of the activation maps. However, these works are difficult to implement and are non-trivial to scale to a wide variety of tasks in challenging benchmarks.

Another disadvantage of using voxel grids is that they build on the assumption that the scene has an Euclidean structure and is not invariant to transformations such as isotropy and non-rigid deformations. This limitation can be overcome by considering the points as members of an orderless set, which are used along a global representation~\cite{qi2016pointnet, qi2017pointnetpp, zaheer2017deep} of the 3D volume.


Approaches such as~\cite{kamnitsas2017efficient, tchapmi2017segcloud} have used a CRF for post-processing the semantic segmentation prediction from a 3D ConvNet. In~\cite{qi20173d}, the authors used a 3D graph neural network to iteratively improve the unary semantic segmentation predictions. Our approach can be used to provide better a unary term for these methods.

\paragraph{3D Surface based Descriptors.} A different approach would be to reason along the surface of the 3D volumes. \cite{johnson1999using} introduced the idea of spin images, which builds a 3D surface based descriptors for object recognition. \cite{huang2015analysis} learns a generative model to produce object structure through surface geometry. \cite{masci2015geodesic, boscaini2016learning} extended the idea of convolutions to non-Euclidean structures by learning anisotropic heat kernels which relates to surface deformity. However, such methods require point associations to learn the filters, which are difficult to obtain for range data depicting natural scenes.~\cite{kalogerakis20163d} combines the segmentation results of multiple views into a surface representation of the 3D object. This is followed by a post-processing step with a CRF which smoothens the labels along the surface geometry. Such smoothening CRF can be used to further improve the results of our approach as well.

\paragraph{Multi-Scale 2D Convolutions.} To get better performance, a host of approaches have used multi-scale input to for tasks such as semantic segmentation~\cite{chen2016attention, lin2016efficient}, optical flow~\cite{wang2016autoscaler}, and detection~\cite{hao2017scale, hu2016finding}. Other approaches include adaptively learning pixel-wise scale~\cite{dai2017deformable, zhang2017scale}, and upscaling feature activations to combine multiple scales for the final prediction~\cite{lin2016feature, zhao2016pyramid}. The key difference of such approaches with ours is that these scales are arbitrary and do not utilize the 3D geometry of the scene. 

\section{Method}
\newcommand{\mb}[1]{\mathbf{#1}}
\newcommand{\mrm}[1]{\mathrm{#1}}

An image from a single frame RGBD camera captures only the visible surface of the 3D space. Instead of wasting memory on the entire 3D volume, we introduce SurfConv, which concentrates the computation only along the visible surface. In Sec.~\ref{sec:derivation} we derive SurfConv, which approximates 3D convolution operation to a depth-aware multi-scale 2D convolution. We justify the approximation assumption and its implications in Sec.~\ref{sec:discussion}. 
In Sec.~\ref{sec:d4}, we describe the $D^4$ scheme that determines the scales in a systematic fashion. 
	
\subsection{Surface Convolution}
\label{sec:derivation}
\textbf{Notation}. We denote a point detected by the sensor as $\mb{p}$. Three scalars ($\mrm{p_x}$,$\mrm{p_y}$,$\mrm{p_z}$) represents its position in 3D. Following the classic camera model, we set the sensor position at the origin, and the principal axis as the positive $\mb{z}$ direction. The distance from the image plane to the camera center is the camera's focal length. For simplicity, we rescale the coordinates such that the focal length is equal to 1. We can then compute the image plane coordinates of point $\mb{p}$ using standard perspective projection as:

\begin{equation} \label{eq:proj}
\begin{aligned}
\mrm{p_i}=\frac{\mrm{p_x}}{\mrm{p_z}}, \quad
\mrm{p_j}=\frac{\mrm{p_y}}{\mrm{p_z}}
\end{aligned}
\end{equation}

We denote the information (e.g. color intensity values) of point $\mb{p}$ as $\mb{I_p}$, and its semantic class label as $\mb{L_p}$. At high level, semantic segmentation can be formulated as

\begin{equation} \label{eq:semseg}
\begin{aligned}
\mb{L_p}=\mb{F}(\mb{I}_{\mb{p}'\in \mb{R(p)}})
\end{aligned}
\end{equation}
where $\mb{F}$ is a function of choice (e.g. a convolutional neural network), and $\mb{R(p)}$ defines a local neighborhood of point $\mb{p}$. We refer to $\mb{R(p)}$ as the \textit{receptive field} at point $\mb{p}$, as commonly used in the literature. Different types of convolution take different forms of $\mb{F}$ and $\mb{R}$. Next, we mainly focus on the receptive field $\mb{R}$, as it defines the local neighborhood that affects the final segmentation prediction.

\textbf{Image Convolution}. In image convolution, the receptive field of point $\mb{p}$ is defined as $\mb{R}_{img}(\mb{p})=\{\mb{p}'\}$, such that

\begin{equation}
\begin{aligned}
& \mrm{p_i}'\in[\mrm{p_i}-\mrm{\Delta}_{img},\mrm{p_i}+\mrm{\Delta}_{img}]\\
& \mrm{p_j}'\in[\mrm{p_j}-\mrm{\Delta}_{img},\mrm{p_j}+\mrm{\Delta}_{img}]
\end{aligned}
\end{equation}
where $\mrm{\Delta}_{img}$ defines the receptive field radius. We can see that $\mb{R}_{img}(\mb{p})$ defines a rectangle on the projected image plane. The receptive field has the same number of pixels regardless of the center point's distance. Therefore, image convolution suffers from the scale variance problem.

\textbf{3D Convolution}. To utilize the 3D information, especially in the depth dimension, 3D convolution has been introduced. In 3D convolution, the receptive field can be defined by trivially extending into all three spatial dimensions, i.e. $\mb{R}_{3d}(\mb{p})=\{\mb{p}'\}$ such that

\begin{equation} \label{eq:3d}
\begin{aligned}
& \mrm{p_x}'\in[\mrm{p_x}-\mrm{\Delta}_{3d},\mrm{p_x}+\mrm{\Delta}_{3d}]\\
& \mrm{p_y}'\in[\mrm{p_y}-\mrm{\Delta}_{3d},\mrm{p_y}+\mrm{\Delta}_{3d}]\\
& \mrm{p_z}'\in[\mrm{p_z}-\mrm{\Delta}_{3d},\mrm{p_z}+\mrm{\Delta}_{3d}]
\end{aligned}
\end{equation}
This defines a 3D cuboid centered at point $\mb{p}$, with radius $\mrm{\Delta}_{3d}$. In 3D convolution, the receptive field becomes independent to depth and no longer suffers from scale variance. However, for a single-frame 3D sensor, the actual 3D data is essentially a surface back-projected from the image plane. This means at any given 3D cubic receptive field, the majority of space is empty, which makes training $\mb{F}$ difficult. To address the sparsity problem, approaches have used Truncated Signed Distance Function (TSDF)~\cite{zhang2016deepcontext} and flipped-TSDF~\cite{song2016semantic} that fills the empty space, or decrease the voxel resolution~\cite{engelcke2017vote3deep}.

\textbf{Local Planarity Assumption}. We seek a solution that directly operates along the visible surface, where the meaningful information resides. To achieve this, we first introduce the \textit{local planarity} assumption. Then we show that under this assumption, we can reformulate 3D convolution as a depth-aware multi-scale 2D convolution. We name this reformulated approximation as Surface Convolution (SurfConv).

\begin{figure}[t!]
	\begin{center}
		\includegraphics[width=0.92\linewidth]{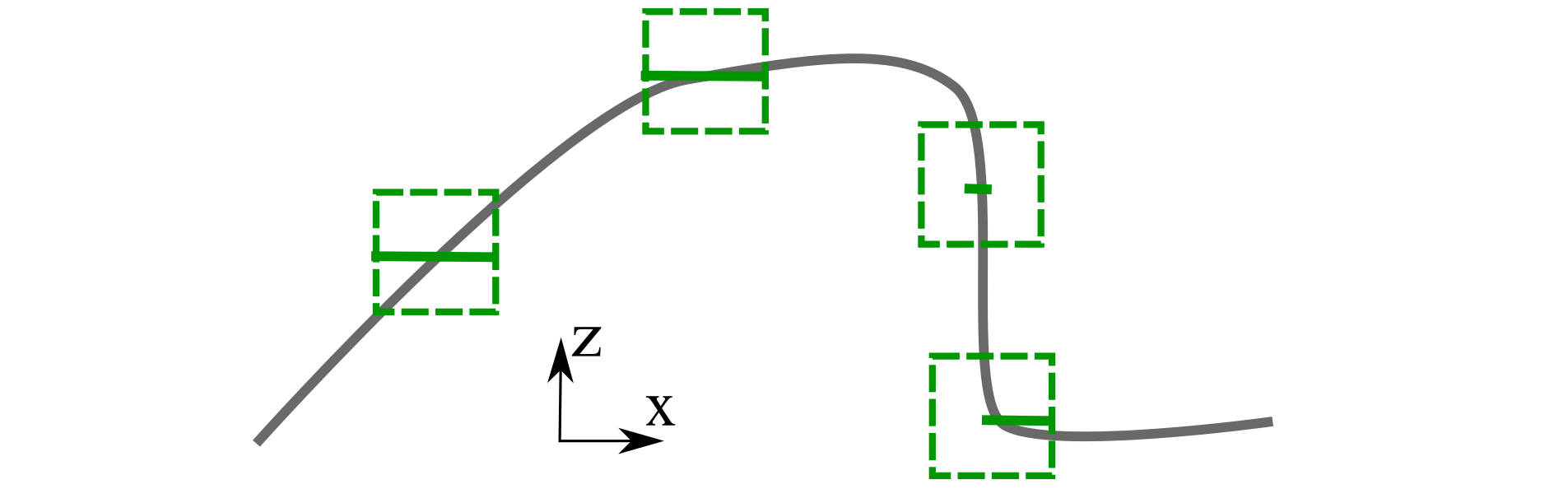}
	\end{center}
	\caption{\footnotesize A 2D illustration of our local planarity assumption. Gray curve shows the visible surface, solid green line shows the approximation plane, for the center point's local neighborhood.}
	\label{fig:plane}
	\vspace{-0.72cm}
\end{figure} 

The local planarity assumption is defined as: \textit{All neighbor points are approximated to have the same depth as the receptive field center}. Fig.~\ref{fig:plane} illustrates the approximation assumption. Under this assumption, we have

\begin{equation} \label{eq:plane}
\begin{aligned}
\mrm{p_z}'=\mrm{p_z}, \quad \forall \mb{p}'\in \mb{R}_{3d}(\mb{p})
\end{aligned}
\end{equation}

\textbf{Surface Convolution}. Under the local planarity assumption, we can transform the 3D convolution receptive field into the SurfConv receptive field. Combining Eq.~\ref{eq:proj} and Eq.~\ref{eq:3d}, we get

\begin{equation}
\begin{aligned}
\mrm{p_i}'\mrm{p_z}'=\mrm{p_x}'\in [\mrm{p_x}-\mrm{\Delta}_{3d},\mrm{p_x}+\mrm{\Delta}_{3d}]
\end{aligned}
\end{equation}
Then we apply the local planarity assumption as in Eq.~\ref{eq:plane}, and get

\begin{equation} \label{eq:plane}
\begin{aligned}
\mrm{p_i}'\in [\frac{\mrm{p_x}-\mrm{\Delta}_{3d}}{\mrm{p_z}},\frac{\mrm{p_x}+\mrm{\Delta}_{3d}}{\mrm{p_z}}]
\end{aligned}
\end{equation}
We can further apply the projection matrix, and obtain the final receptive field definition of SurfConv: $\mb{R}_{sf}(\mb{p})=\{\mb{p}'\}$ such that

\begin{equation} \label{eq:surf}
\begin{aligned}
& \mrm{p_i}'\in[\mrm{p_i}-\frac{\mrm{\Delta}_{sf}}{\mrm{p_z}},\mrm{p_i}+\frac{\mrm{\Delta}_{sf}}{\mrm{p_z}}]\\
& \mrm{p_j}'\in[\mrm{p_j}-\frac{\mrm{\Delta}_{sf}}{\mrm{p_z}},\mrm{p_j}+\frac{\mrm{\Delta}_{sf}}{\mrm{p_z}}]
\end{aligned}
\end{equation}
where $\mrm{\Delta}_{sf}=\mrm{\Delta}_{3d}$ defines the receptive field radius in the 3D space. In this way, the SurfConv receptive field defines a square image region, whose size is controlled by the center point's depth. This means SurfConv is essentially a depth aware multi-scale 2D convolution. This bridges the 3D and 2D perspectives, and avoids the disadvantages of either method. Compared to 2D convolution, SurfConv utilizes 3D data and does not suffer from scale variance. Compared to 3D convolution, SurfConv not only saves the preprocessing step of filling empty voxels, but also enables learning compact, parameter-efficient convolution 2D filters that directly targets the real-world scale of the input data.

\begin{figure}[t!]
	\begin{center}
		\setlength{\tabcolsep}{0pt}
		\begin{tabular}{cc}
			\includegraphics[width=0.46\linewidth]{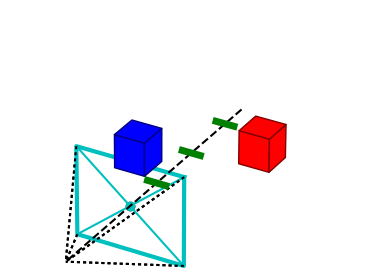} &
			\includegraphics[width=0.46\linewidth]{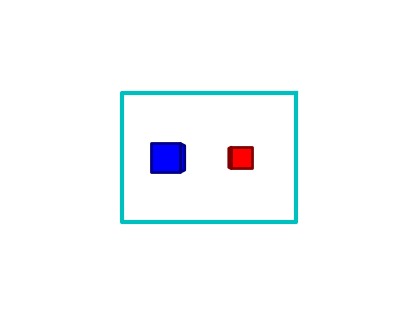}\\
			\multicolumn{2}{c}{{original camera}}\\
			\includegraphics[width=0.46\linewidth]{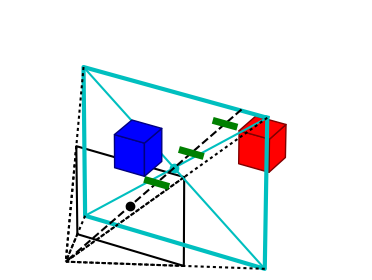} &
			\includegraphics[width=0.46\linewidth]{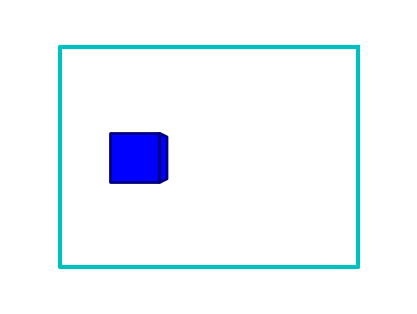}\\
			\multicolumn{2}{c}{{SurfConv at $\mb{z}_1$}}\\
			\includegraphics[width=0.46\linewidth]{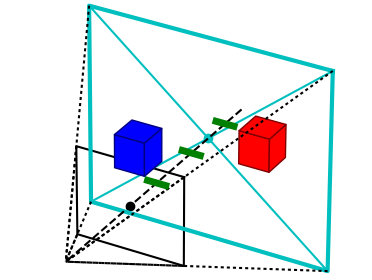} &
			\includegraphics[width=0.46\linewidth]{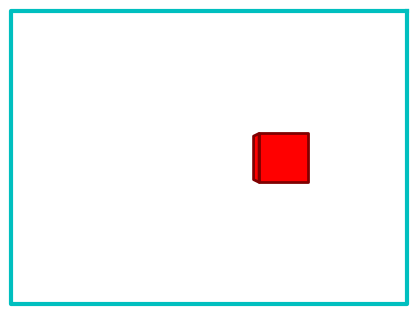}\\
			\multicolumn{2}{c}{{SurfConv at $\mb{z}_2$}}\\
		\end{tabular}
	\end{center}
	\caption{\footnotesize A toy example of discretized SurfConv. In each row, left side shows the scene and projection plane in cyan, right side shows the image. Green bars show boundaries separating discretization levels. At each SurfConv level, only points that have depth within the level are visible. The blue and red objects remain equal-sized to SurfConv, despite their different depth.}
	\label{fig:disc}
	\vspace{-0.4cm}
\end{figure}

In SurfConv, $\mrm{p_z}\in \mathbb{R}$ is a continous variable. This means for each point, we need to dynamically resize the receptive field based on its size determined by Eq.~\ref{eq:surf}, before passing it to the recognition module $\mb{F}$ that takes fixed size input. This is computationally inefficient in practice. To address this problem, we further replace the continues depth with a set of discretized values, i.e. $\mrm{p_z}\in \{\mrm{z_1},\mrm{z_2},...\mrm{z}_\mathcal{N}\}$. We refer to $\mathcal{N}$ as the \textit{level} of SurfConv. With the discretized depth, we can cache $\mathcal{N}$ levels of the image pyramid. Note that since we are interested in surface convolution, each pixel in the original RGBD image belongs to exactly one level of the pyramid. Fig.~\ref{fig:disc} shows a toy example of our discretization process.

\subsection{Bridging 3D and 2D Convolution}
\label{sec:discussion}
In SurfConv, we discretize the $\mb{z}$ dimension into $ \mathcal{N} $ levels and maintain the full resolution in $\mb{x}$ and $\mb{y}$ dimensions. Thus, our surface convolution can be seen as a deformed version of general 3D convolution, where SurfConv has coarser $\mb{z}$ resolution consisting of $\mathcal{N}$ levels, and divides the 3D space into a $\mb{z}$-stretched voxel grid. 
The memory constraints of current day GPUs limits the resolution of the input. 
In 3D convolution, the 3D space is discretized similarly in all three axes. This results in large grids and lowered maximum feasible resolution. In contrast, SurfConv maintains the full resolution along axes parallel to the image projection plane ($ \mb{x} $ and $ \mb{y} $), and have a much coarser resolution for the axis perpendicular to the image plane (i.e. $ \mb{z} $). In an RGBD image, the information only resides along the visible surface. This motivates the lower $\mb{z}$ resolution, because information is scarce along this direction. 
Practically, SurfConv can be simply implemented with a depth-aware multi-scale 2D convolution. Each depth level consists a proportionally scaled version of the input, masked to contain points within its depth range. Standard 2D CNN training is applied to all levels simultaneously. Therefore, SurfConv can easily benefit from networks pre-trained on a variety of large-scale 2D image datasets. 

\begin{figure}[t!]
	\begin{center}
		\includegraphics[width=0.92\linewidth]{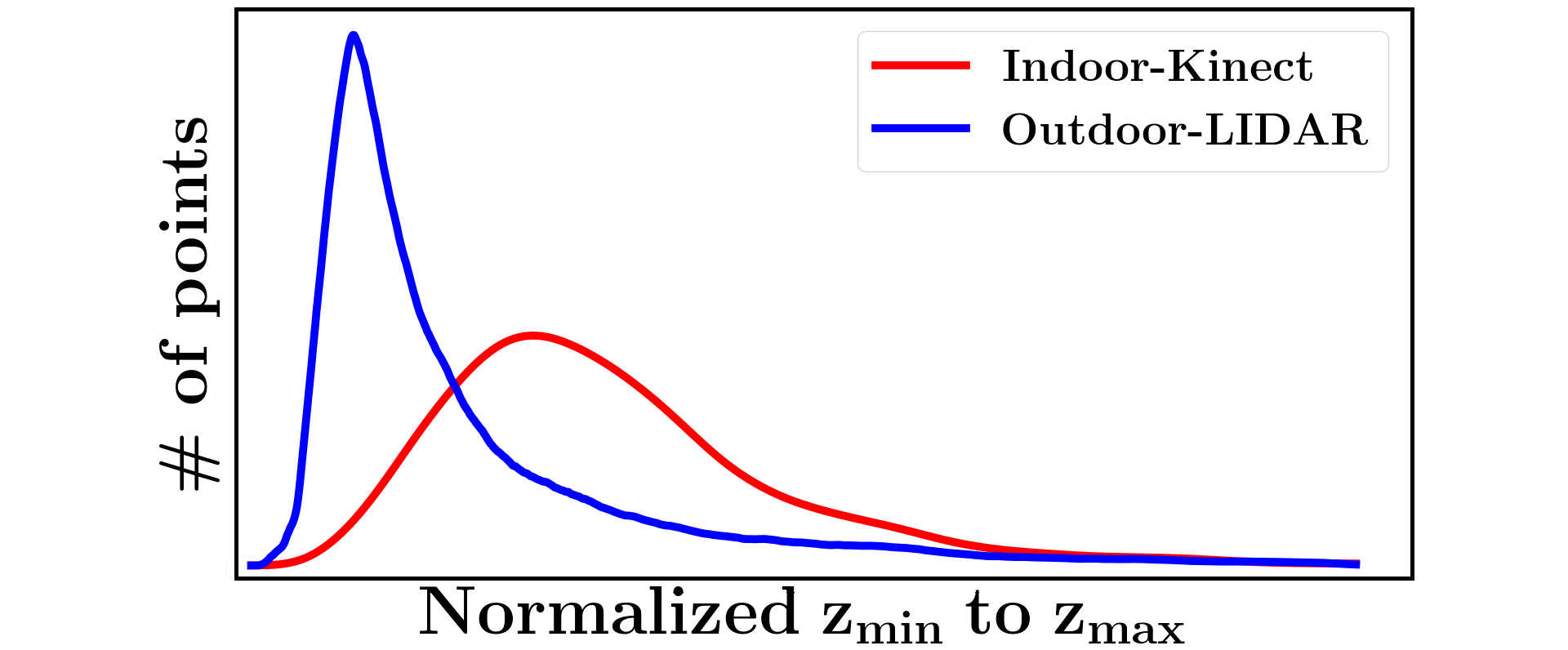}
	\end{center}
	\caption{\footnotesize Point depth distribution at different environment. The distribution leans heavily to the lower side due to occlusion and sensor resolution. Statistics obtained from NYUv2~\cite{silberman2012indoor} and KITTI~\cite{geiger2013vision}.}
	\label{fig:depth}
	\vspace{-0.4cm}
\end{figure}

\subsection{$D^4$: Data-Driven Depth Discretization}
\label{sec:d4}

To obtain a set of discretized depth levels, uniform bins are sub-optimal. This is because in single-viewpoint input data, near points significantly out-number far points due to occlusion and decreasing resolution over depth. Fig.~\ref{fig:depth} shows actual depth distributions from real indoor and outdoor data. Therefore, uniform bins result in unbalanced data allocation between levels, i.e. the first few levels have almost all points, while the last few levels are almost empty.

To address this problem, we introduce the $D^4$ scheme. Instead of dividing levels evenly, we compute level boundaries such that all levels contribute the same amount of influence to the segmentation model $\mb{F}$. First, we define the \textit{importance function} of a point as 

\begin{equation} \label{eq:surf}
\begin{aligned}
\mb{\Theta}(\mb{p})=\mrm{p_z}^{\gamma}
\end{aligned}
\end{equation}
where we refer to $\gamma$ as the \textit{importance index}. We use the importance function to assign a weight to each input point, then we find $\mathcal{N}$ discretization levels such that all levels possess the same amounts summed importance.

Intuitively, with $\gamma=0$, all points are equally important regardless of their depth. As result, all levels are allocated with same number of image pixels. With $\gamma=2$, a point's influence is proportional to the back-projected 3D surface area it covers. As result, all levels have equal amount of total 3D surface area after the discretization and allocation.

Ideally, $\gamma=2$ seems the optimal setting because it divides the visible surface area evenly to different levels. However, we argue that $\gamma$ should instead be a hyper-parameter. Data quality decreases over distance for sensors, i.e. the farther an object is, the less detailed measurement a sensor receives. A farther object occupies a smaller field of view from the sensor's viewpoint. This means lower resolution, hence lower capture quality. Additionally, in sensors such as stereo cameras and Microsoft Kinect, precision decreases as depth increases, making farther points inherently more noisy. Therefore, in order to learn the best recognition model, there exists a tradeoff between trusting near clear data, and paying attention to adapt to far noisy data. In other words, the best index is determined by $\hat{\gamma}=2-\zeta$, where $\zeta$ quantifies this near-far tradeoff. It is difficult to analytically compute $\zeta$, because it depends on the actual sensor configuration and scene properties. Therefore, we tune $\zeta$, hence $\gamma$, through validation on the actual data.

\section{Experimental Results}

\newcommand{\fir}[1]{\textbf{\textcolor{red}{#1}}}
\newcommand{\sece}[1]{\textbf{\textcolor{blue}{#1}}}

We demonstrate the effectiveness of our approach by showing results on two real-world datasets (KITTI~\cite{geiger2013vision} and NYUv2~\cite{silberman2012indoor}) for the 3D semantic segmentation task.

\subsection{Experimental Setup}
\paragraph{CNN model.} We use the skip-connected fully convolutional architecture~\cite{long2015fully} with two different backbones:
\begin{enumerate}
	\item \textit{ResNet-18}~\cite{he2016deep}: We modify the size of all convolutional kernels to $ 3 \times 3 $ and experiment with different number of feature channels in each layer. We try light and heavy weight versions where the number of feature channels are $ 1/4 $, $ 1/2 $, same, or twice of the number of original channels. The input to our network is a 6-channel RGB+HHA~\cite{gupta2014learning} image. This network has been trained from scratch, similar to the baseline 3D convolution based approaches. 
	\item \textit{VGG-16}~\cite{simonyan2014very}: We particularly choose this model because it is conventionally used in previous work on the NYUv2 dataset~\cite{long2015fully,gupta2016cross}. The input to this network is the standard RGB image. This model has been pre-trained on the Imagenet dataset.
\end{enumerate}

Using the light weight models, we show that our performance is competitive (NYUv2) or better (KITTI) than the state-of-the-art 3D convolution based approaches even with about a quarter of their parameters. Since the memory requirement of our network is low compared to 3D convolution based approaches, we can take advantage of heavier models to further improve our performance. Moreover, our approach can take advantage of pre-trained weights on existing large scale 2D datasets. For training our networks, we follow FCN-8s and use the logarithm loss function.

\begin{table*}[t!]
	\centering
	\begin{tabular}{l|cc|cc|cccc}
~ &  3d & RGB & \# of para. & infer./ms & $IOU_{img}$ & $Acc_{img}$ & $IOU_{surf}$ & $Acc_{surf}$\\ 
\hline
\multirow{3}{*}{Conv3D~\cite{tchapmi2017segcloud,song2016semantic}} & ftsdf & no & 233k & \sece{8} & 12.43 & 50.05 & 12.69 & 53.34\\
~ & no & yes & 238k & 10 & 12.36 & 48.44 & 12.66 & 51.29\\
~ & ftsdf & yes & 241k & 11 & \fir{13.19} & 49.85 & \fir{13.65} & 52.89\\
\hline
\multirow{3}{*}{PointNet~\cite{qi2016pointnet}} & xyz & no & 1675k & 118 & 6.25 & 46.44 & 5.82 & 47.46\\
~ & xyz-G & no & 1675k & 118 & 6.54 & 46.88 & 6.16 & 47.85\\
~ & xyz-G & yes & 1675k & 117 & 6.87 & 47.35 & 6.47 & 48.21\\
\hline
DeformCNN~\cite{dai2017deformable} & HHA & yes & 101k & 6 & 12.82 & \fir{55.05} & 11.67 & 54.12\\
\hline
\textit{SurfConv}1 & HHA & yes & \fir{65k} & \fir{5} & 12.31 & \sece{53.74} & 11.27 & 54.24\\
\textit{SurfConv}4-$\gamma$1.0 & HHA & yes & \fir{65k} & 26 & 12.01 & 52.19 & 11.98 & \sece{55.44}\\
\textit{SurfConv}4-$\gamma$2.0 & HHA & yes & \fir{65k} & 24 & \sece{13.10} & 53.48 & \sece{12.79} & \fir{55.99}\\
	\end{tabular}
	\caption{\small Training different models from scratch on NYUv2~\cite{silberman2012indoor}. All models are trained till convergence for five times, and average performance is reported. All training are performed in a data-augmentation-free fashion, but thorough searching in the training hyper-parameter space is guaranteed. We mark the \fir{best} and \sece{second best} method in blue and red. Compared to Conv3D~\cite{tchapmi2017segcloud,song2016semantic}, SurfConv achieves close IOU performance and better Acc performance, while using 30\% number of parameters. Compared to PointNet~\cite{qi2016pointnet}, SurfConv achieves 6\% improvement across all measures, while only using less than 5\% number of parameters. Compared to DeformCNN~\cite{dai2017deformable} SurfConv achieves better or close measurements with 64\% number of parameters. Furthermore, when pre-training with ImageNet, SurfConv achieves a huge boost in performance (10\% improvement in all metrics as shown in Fig.~\ref{fig:nyu_finetune}).}
	\label{tab:nyu_scratch}
	\vspace{-0.36cm}
\end{table*}

\vspace{-0.36cm}
\paragraph{Baselines.}  We compare our approach with Conv3D~\cite{tchapmi2017segcloud,song2016semantic}, PointNet~\cite{qi2016pointnet}, and DeformCNN~\cite{dai2017deformable}. For Conv3D, we use the SSCNet architechture~\cite{song2016semantic}, and train it with three variations of gravity-aligned voxel input: RGB, flipped-TSDF, and both. We follow~\cite{song2016semantic} and use the maximum possible voxel resolution that can fit a single-sample batch into 12GB memory, which results in a 240$\times$144$\times$240 voxel grid (with 2\textit{cm} resolution) on NYUv2, and a 400$\times$60$\times$320 voxel grid (with 10\textit{cm} resolution) on KITTI. The points that fall into the same voxel are given the same predicted label in inference. 

For PointNet, we directly use the published source code, and train it on three types of input: original point cloud, gravity-algined point cloud, and RGB plus gravity alignment. We randomly sample points from the point cloud as suggested in the paper. Specifically, we set the sample number as 25K, which fills 12GB memory with batch size 8.

For DeformCNN, we replace \texttt{res5} layers with deformable convolution as recommended in~\cite{dai2017deformable}. We try jointly training all layers of DeformCNN, as well as training with deformation offset frozen before the joint training. We report measurements of the latter for its better performance. For fair comparison, we further augment DeformCNN to use depth information by adding extra HHA channels.

SurfConv with a single level is equivalent to the FCN-8s~\cite{long2015fully} baseline. All models are trained using the original data as-is, without any augmentation tricks.

\vspace{-0.36cm}
\paragraph{Metrics.} For all experiments, we use the pixel-wise accuracy ($ Acc $) and the intersection over union ($ IOU $) metrics. We report  these metrics on both pixel-level ($Acc_{img}$ and $IOU_{img}$) and surface-level ($Acc_{surf}$ and $IOU_{surf}$). For the surface level metrics, we weigh each point by its surface area in 3D to compute the metrics. To reduce model sensitivity to initialization and random shuffling order in training, we repeat all experiments five times on a Nvidia TitianX GPU, and report the average model performance.



\subsection{NYUv2}
NYUv2~\cite{silberman2012indoor} is a semantically labeled indoor RGB-D dataset captured by a Kinect camera. In this dataset, we use the standard split of 795 training images and 654 testing images. We randomly sample 20\% rooms from the training set as the validation set. The hyper-parameters are chosen based on the best mean IOU on the validation set, which we then use to evaluate all metrics on the test set. For the label space, we use the 37-class setting~\cite{gupta2014learning,qi20173d}. To obtain 3D data, we use the hole-filled dense depth map provided by the dataset. Training our model over all repetitions and hyper-parameters takes a total of 950 GPU hours.

\begin{figure}[t!]
\begin{center}
\setlength{\tabcolsep}{0pt}
\begin{tabular}{cc}
\includegraphics[width=0.47\linewidth]{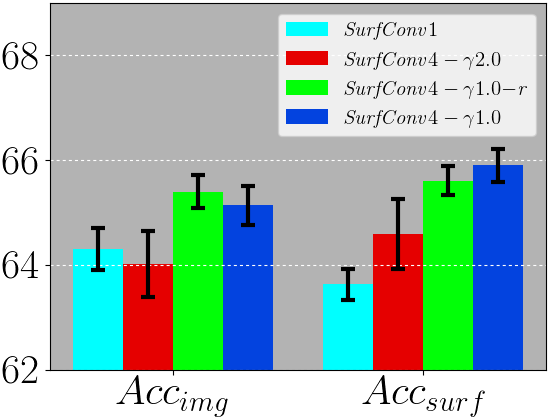}&
\includegraphics[width=0.47\linewidth]{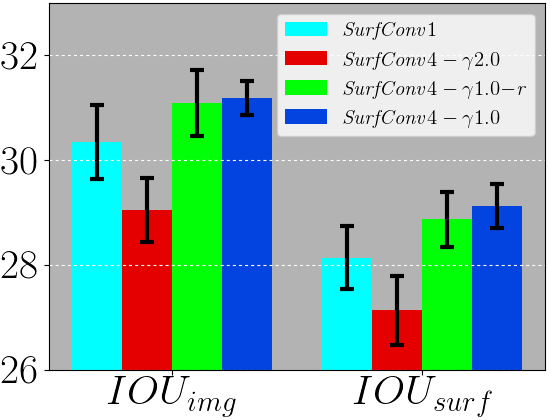}\\
\end{tabular}
\end{center}
\vspace{-0.36cm}
\caption{\footnotesize Mean performance and standard deviation of NYUv2 finetuning. Comparing to the vanilla CNN model (i.e. SurfConv1), 4-level SurfConv is able to improve on both image-wise and surface-wise metrics. $r$ denotes the reweighted version.}
\label{fig:nyu_finetune}
\vspace{-0.36cm}
\end{figure}

The result is shown in Table~\ref{tab:nyu_scratch}. Compared to Conv3D, SurfConv achieves close performance on IOU and better performance on accuracy, while using 30\% of its number of parameters. Compared to PointNet, SurfConv achieves 6\% improvement across all metrics, while only using less than 5\% of its number of parameters.  Compared to the latest scale-adaptive architecture DeformCNN, SurfConv is more suitable for RGBD images because it uses depth information more effectively, achieving better or close performance while using fewer parameters. Having more number of weights (VGG-16 architecture) and pre-training with Imagenet gives us a huge boost in performance (Fig.~\ref{fig:nyu_finetune}).

\begin{figure*}[t!]
\begin{center}
\setlength{\tabcolsep}{0pt}
\begin{tabular}{cc}
\includegraphics[width=0.6838\linewidth]{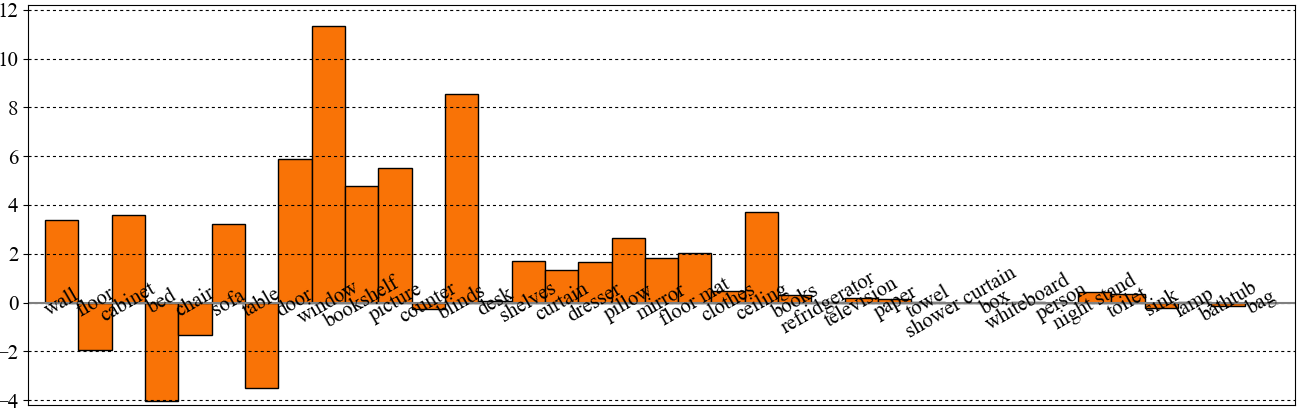}&
\includegraphics[width=0.2362\linewidth]{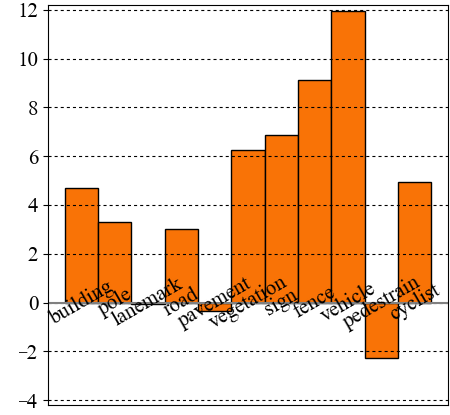}\\
NYUv2-37class & KITTI-11class
\end{tabular}
\end{center}
\vspace{-0.36cm}
\caption{\footnotesize Average improved percentage of per-class surface IOU, using multi-level SurfConv over the single-level baseline, with the exact same CNN model $\mb{F}$ (Eq.~\ref{eq:semseg}). Models are trained from scratch. On NYUv2, we improve 27/37 classes with 1.40\% mean IOU increasement. On KITTI, we improve 8/11 classes with 4.31\% mean IOU increasement.}
\label{fig:class}
\vspace{-0.36cm}
\end{figure*}

\begin{figure*}[t!]
\begin{center}
\setlength{\tabcolsep}{0pt}
\begin{tabular}{cc}
\includegraphics[width=0.6838\linewidth]{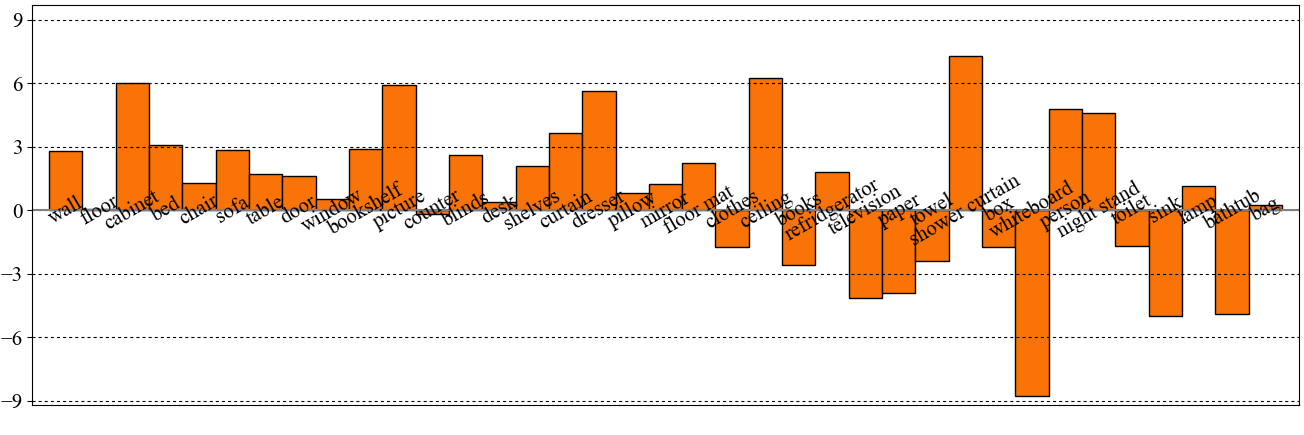}&
\includegraphics[width=0.2362\linewidth]{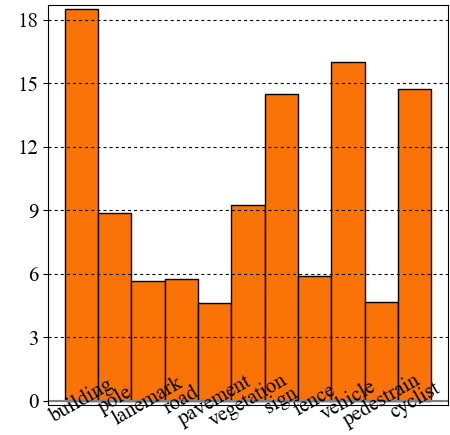}\\
NYUv2-37class & KITTI-11class
\end{tabular}
\end{center}
\vspace{-0.36cm}
\caption{\footnotesize Same as Fig.~\ref{fig:class}, but finetuning from ImageNet instead of training from scratch. On NYUv2, multi-level SurfConv improves 26/37 classes with 0.99\% mean IOU increasement, from the single-level baseline. On KITTI, multi-level SurfConv improves 11/11 classes with 9.86\% mean IOU increasement.}
\label{fig:class_ft}
\vspace{-0.36cm}
\end{figure*}

Comparing SurfConv with different levels trained from scratch in Table~\ref{tab:nyu_scratch}, it can be seen that the 4-level model is slightly better or close to the 1-level model in image-wise metrics, and significantly better in surface-wise metrics. Using pre-trained network (Fig.~\ref{fig:nyu_finetune}),  our 4-level SurfConv achieves better performance than the vanilla single-level model (FCN-8s~\cite{long2015fully} baseline), especially in the surface-wise metrics.  We also explore a SurfConv variant where the training loss for each point is re-weighted by its area of image-plane projection, marked by $r$. This makes the training objective closer to $Acc_{img}$. The re-weighted version achieves slightly better image-wise performance, at the cost of having slightly worse surface-wise performance.

\subsection{KITTI}
KITTI~\cite{geiger2013vision} provides parallel camera and LIDAR data for outdoor driving scenes. We use the semantic segmentation annotation provided in~\cite{xu2016multimodal}, which contains 70 training and 37 testing images from different scenes, with high quality pixel annotations in 11 categories. Due to the smaller dataset size and lack of standard validation split, we directly validate all compared methods on the held-out testing set. To obtain dense points from sparse LIDAR input, we use a simple real-time surface completion method that exhaustively join adjacent points into mesh triangles. The densified points are used as input for all methods evaluated. The smaller size of KITTI allows us to thoroughly explore different settings of SurfConv levels, influence index $\gamma$, as well as CNN model capacity. Our KITTI experiments take a total of 750 GPU hours.

\begin{figure*}[t!]
	\begin{center}
		\setlength{\tabcolsep}{0pt}
		\begin{tabular}{cccc}
			\includegraphics[width=0.23\linewidth]{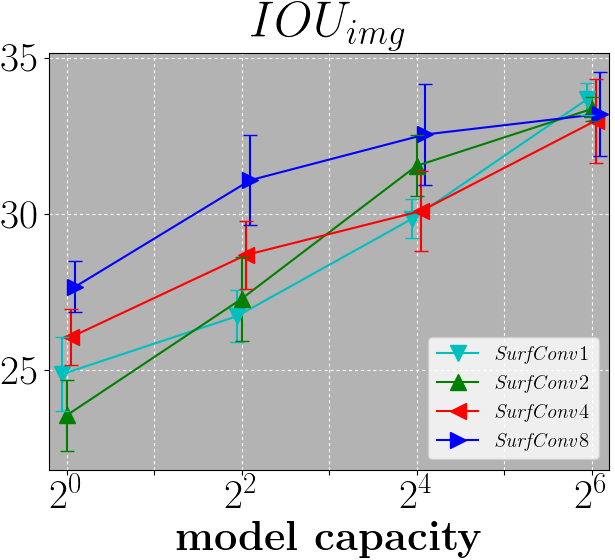}&
			\includegraphics[width=0.23\linewidth]{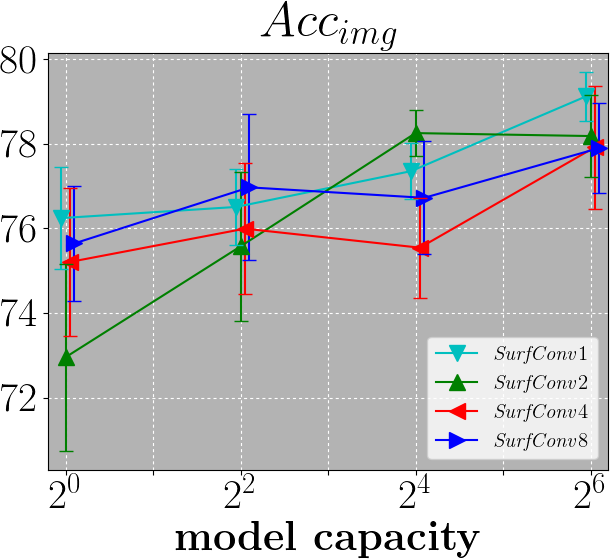}&
			\includegraphics[width=0.23\linewidth]{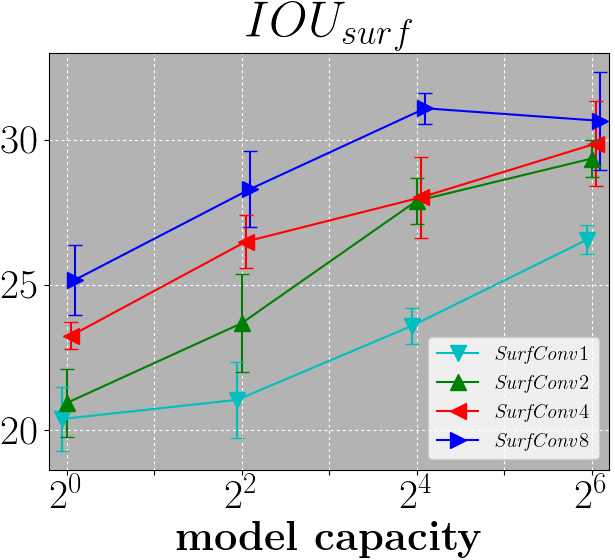}&
			\includegraphics[width=0.23\linewidth]{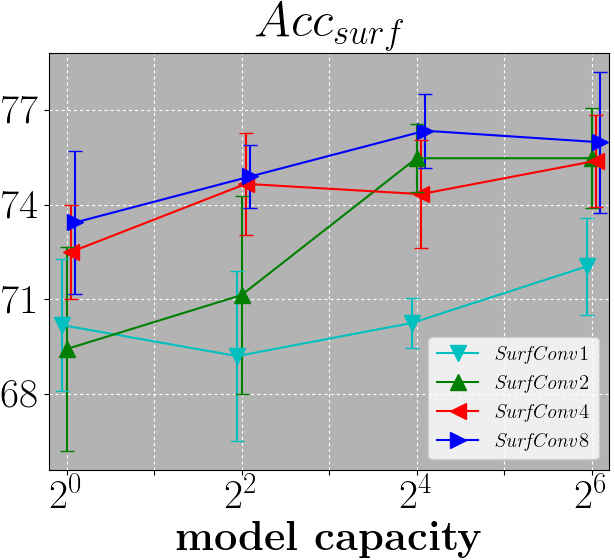}\\
		\end{tabular}
	\end{center}
	\vspace{-0.36cm}
	\caption{\footnotesize Exploring the effect of model capacity with different SurfConv levels, on the KITTI dataset. Using exactly the same model ($\mb{F}$ in Eq.~\ref{eq:semseg}), multi-level SurfConv achieves significantly better surface-wise performance, while maintaining better or similar image-wise performance. All models are trained from scratch using $\gamma=1$ for five times. Base level of model capacty (i.e. $2^0$) has 65k parameters.}
	\label{fig:kitti_cap}
	\vspace{-0.36cm}
\end{figure*}

\begin{figure*}[t!]
	\begin{center}
		\setlength{\tabcolsep}{0pt}
		\begin{tabular}{cccc}
			\includegraphics[width=0.23\linewidth]{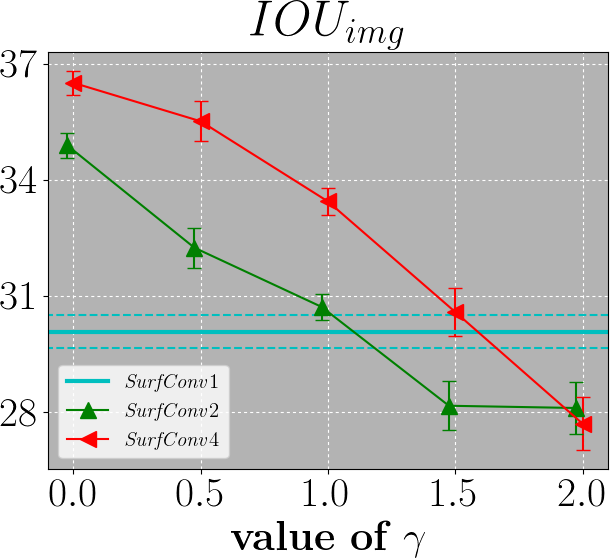}&
			\includegraphics[width=0.23\linewidth]{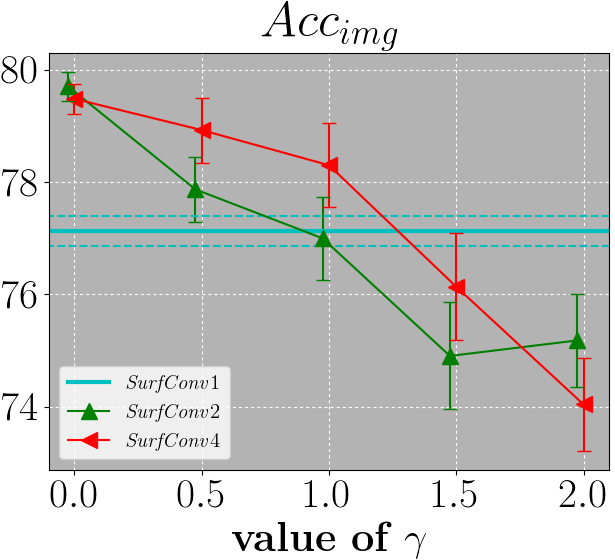}&
			\includegraphics[width=0.23\linewidth]{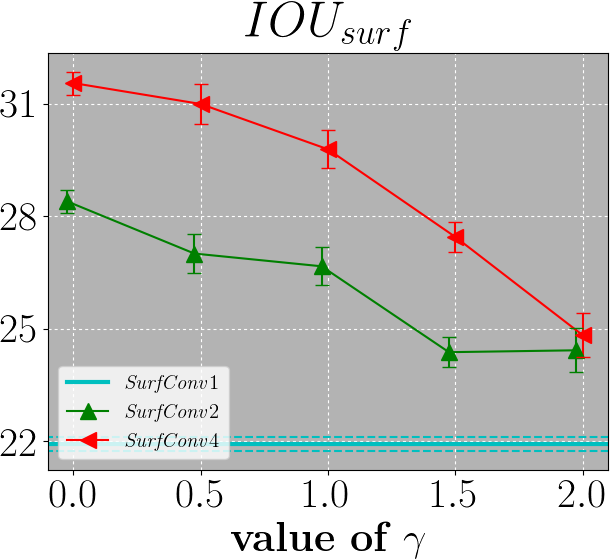}&
			\includegraphics[width=0.23\linewidth]{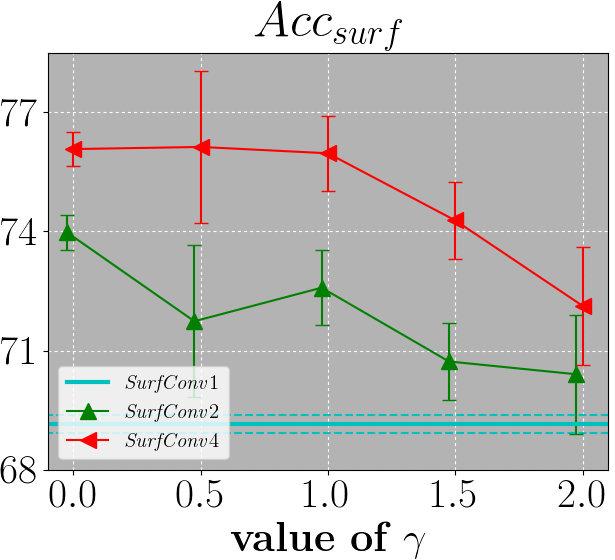}\\
		\end{tabular}
	\end{center}
	\vspace{-0.36cm}
	\caption{\footnotesize Finetuning from an ImageNet pre-trained CNN using different importance index value $\gamma$ and different SurfConv levels, on the KITTI dataset. All models are trained five times. Only three RGB channels are used in this experiment.}
	\label{fig:kitti_ft}
	\vspace{-0.36cm}
\end{figure*}

\begin{figure*}[t!]
	\begin{center}
		\setlength{\tabcolsep}{0pt}
		\begin{tabular}{cccc}
			\includegraphics[width=0.23\linewidth]{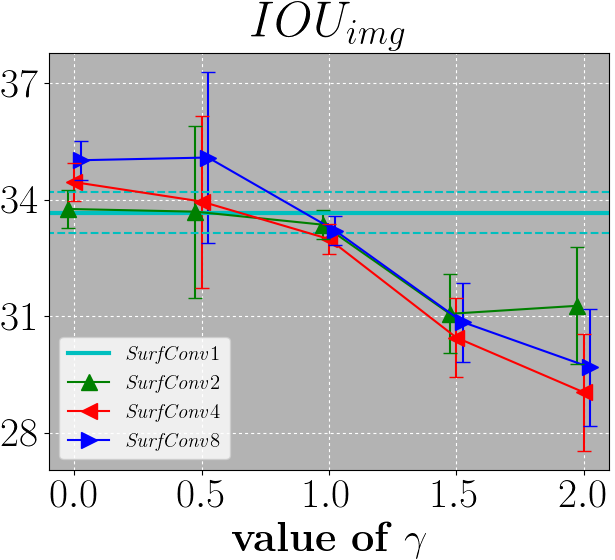}&
			\includegraphics[width=0.23\linewidth]{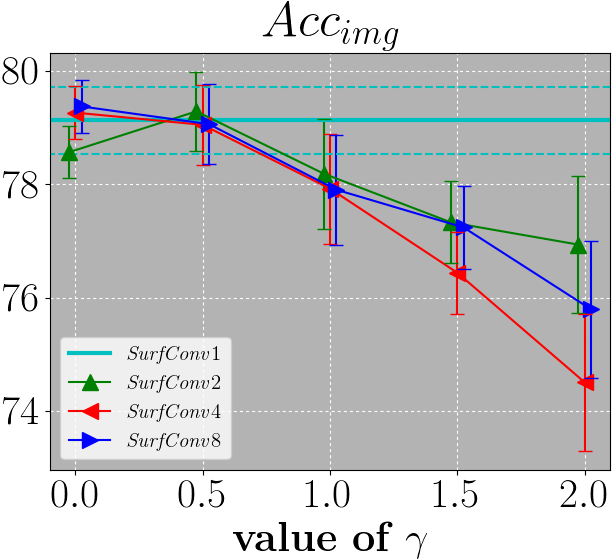}&
			\includegraphics[width=0.23\linewidth]{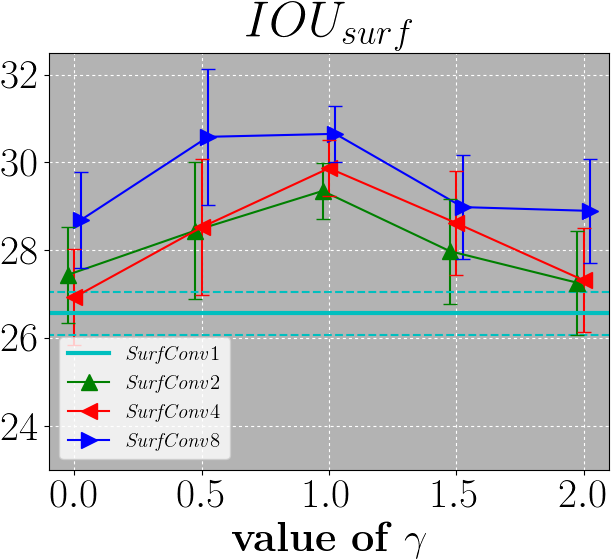}&
			\includegraphics[width=0.23\linewidth]{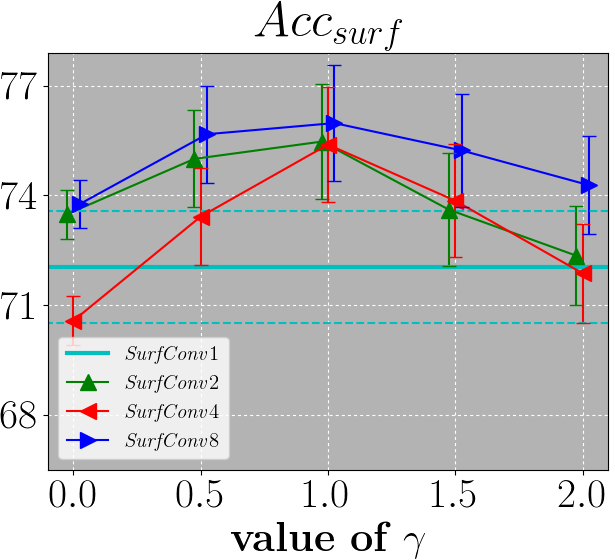}\\
		\end{tabular}
	\end{center}
	\vspace{-0.36cm}
	\caption{\footnotesize Exploring the effect of $\gamma$ when training different levels of SurfConv from scratch. All models are trained five times with capacity $2^6\times$65k.}
	\label{fig:kitti_gamma}
\end{figure*}

\vspace{-0.36cm}
\paragraph{Baseline comparisons.} Table~\ref{tab:kitti_scratch} lists the comparison with baseline methods. SurfConv outperforms all comparisons in all metrics. In KITTI, the median maximum scene depth is 75.87\textit{m}. This scenario is particularly difficult for Conv3D, because voxelizing the scene with sufficient resolution would result in large tensors and makes training Conv3D difficult. On the contrary, SurfConv can be easily trained because its compact 2D filters do not suffer from insufficient memory budget. DeformCNN performs better than image convolution (i.e. \textit{SurfConv}1) for its deformation layers that adapts to object scale variance. However, multi-level SurfConv achieves more significant improvement, demonstrating its capablity of using RGBD data more effectively.

\begin{table}[t!]
	\centering
	\setlength{\tabcolsep}{2pt}
	\begin{tabular}{l|llll}
		~ & $IOU_{img}$ & $Acc_{img}$ & $IOU_{surf}$ & $Acc_{surf}$\\ 
		\hline
		Conv3D~\cite{tchapmi2017segcloud,song2016semantic} & 17.53 & 64.54 & 17.38 & 62.58 \\
		PointNet~\cite{qi2016pointnet} & 9.41 & 55.06 & 9.07 & 64.38\\
		DeformCNN~\cite{dai2017deformable} & 34.24 & 79.17 & 27.51 & 73.36\\
		\hline
		\textit{SurfConv}1 & 33.67 & 79.13 & 26.56 & 72.04\\
		\textit{SurfConv}-best & \textbf{35.09} & \textbf{79.37} & \textbf{30.65} & \textbf{75.97}\\
	\end{tabular}
	\caption{\small Training from scratch on KITTI~\cite{geiger2013vision,xu2016multimodal}. All methods are tuned with thorough hyper-parameter searching, then trained five times to obtain average performance.}
	\label{tab:kitti_scratch}
\end{table}

\vspace{-0.36cm}
\paragraph{Model capacity.} We study the effect of CNN model capacity across different SurfConv levels. To change the model capacity, we widen the model by adding more feature channels, while keeping the same number of layers. This results in 4 capacities that has \{$2^0$,$2^2$,$2^4$,$2^6$\}$\times$65k parameters. We empirically set $\gamma =1$ for all models in this experiment. Fig.~\ref{fig:kitti_cap} shows the result. It can be seen that a higher level SurfConv models have better or similar image-wise performance, while being significantly better in surface-wise metrics. In general, the performance increases as SurfConv level increases. This is because higher SurfConv level enables closer approximation to the scene geometry.

\vspace{-0.36cm}
\paragraph{Finetuning.} Similar to our NYUv2 experiment, we compare multi-level SurfConv with the single-level baseline. The relatively smaller dataset size allows us to also thoroughly explore different $\gamma$ values (Fig.~\ref{fig:kitti_ft}). It can be seen that with a good choice of $\gamma$, multi-level SurfConv is able to achieve significant improvement over the single-level baseline in all image-wise and surface-wise metrics, while using exactly the same CNN model ($\mb{F}$ in Eq.~\ref{eq:semseg}). Comparing NYUv2 and KITTI, it can be seen that our improvement on KITTI is more significant. We credit this to the larger depth range of KITTI data, where scale-invariance plays an important role in segmentation success.

\subsection{Influence of $\gamma$}
The influence index $\gamma$ is an important parameter for SurfConv. We therefore further explore its effects. The optimal values of $ \gamma $ can be different depending on whether the model has been trained from scratch or it has been pre-trained, as shown in Table~\ref{tab:nyu_scratch} and Fig.~\ref{fig:nyu_finetune}. On NYUv2, $\gamma = 1$ is better for finetuning and $ \gamma = 2 $ is better for training from scratch. The pre-trained models are adapted to the Imagenet dataset where most objects are clearly visible and close to camera. The $ \gamma = 1 $ setting weighs the farther points less, which results in a larger number of points at the discretized bin with the largest depth value. In this way, the model is forced to spend more effort on low-quality far points. The observation of lower optimal $\gamma$ on pre-trained networks is further verified by our KITTI results, where $\gamma = 0 $ and $ \gamma = 0.5 $ achieve best results for pre-trained and from-scratch networks respectively. In KITTI, good $\gamma$ values are in general lower than in NYUv2. We attribute this to the fact that in KITTI, besides having a larger range of depth values, the peak of the depth distribution (Fig.~\ref{fig:depth}) occurs much earlier. 

\section{Conclusion}
We proposed SurfConv to bridge and avoid the issues with both 3D and 2D convolution on RGBD images. SurfConv was formulated as a simple depth-aware multi-scale 2D convolution, and realized with a Data-Driven Depth Discretization scheme. We demostrated the effectiveness of SurfConv on indoor and outdoor 3D semantic segmentation datasets. SurfConv
achieved state-of-the-art performance while using less than 30\% parameters used by 3D convolution based approaches.

{\small
	\bibliographystyle{ieee}
	\bibliography{egbib}

\begin{thebibliography}{10}\itemsep=-1pt

\bibitem{boscaini2016learning}
D.~Boscaini, J.~Masci, E.~Rodol{\`a}, and M.~Bronstein.
\newblock Learning shape correspondence with anisotropic convolutional neural
  networks.
\newblock In {\em NIPS}, 2016.

\bibitem{chen2016attention}
L.-C. Chen, Y.~Yang, J.~Wang, W.~Xu, and A.~L. Yuille.
\newblock Attention to scale: Scale-aware semantic image segmentation.
\newblock In {\em CVPR}, 2016.

\bibitem{chen20173d}
X.~Chen, K.~Kundu, Y.~Zhu, H.~Ma, S.~Fidler, and R.~Urtasun.
\newblock 3d object proposals using stereo imagery for accurate object class
  detection.
\newblock {\em TPAMI}, 2017.

\bibitem{chen2016multi}
X.~Chen, H.~Ma, J.~Wan, B.~Li, and T.~Xia.
\newblock Multi-view 3d object detection network for autonomous driving.
\newblock In {\em CVPR}, 2017.

\bibitem{dai2017scannet}
A.~Dai, A.~X. Chang, M.~Savva, M.~Halber, T.~Funkhouser, and M.~Nie{\ss}ner.
\newblock Scannet: Richly-annotated 3d reconstructions of indoor scenes.
\newblock {\em CVPR}, 2017.

\bibitem{dai2017deformable}
J.~Dai, H.~Qi, Y.~Xiong, Y.~Li, G.~Zhang, H.~Hu, and Y.~Wei.
\newblock Deformable convolutional networks.
\newblock In {\em ICCV}, 2017.

\bibitem{deng2017amodal}
Z.~Deng and L.~J. Latecki.
\newblock Amodal detection of 3d objects: Inferring 3d bounding boxes from 2d
  ones in rgb-depth images.
\newblock In {\em CVPR}, 2017.

\bibitem{engelcke2017vote3deep}
M.~Engelcke, D.~Rao, D.~Z. Wang, C.~H. Tong, and I.~Posner.
\newblock Vote3deep: Fast object detection in 3d point clouds using efficient
  convolutional neural networks.
\newblock In {\em ICRA}, 2017.

\bibitem{fang20153d}
Y.~Fang, J.~Xie, G.~Dai, M.~Wang, F.~Zhu, T.~Xu, and E.~Wong.
\newblock 3d deep shape descriptor.
\newblock In {\em CVPR}, 2015.

\bibitem{ge20173d}
L.~Ge, H.~Liang, J.~Yuan, and D.~Thalmann.
\newblock 3d convolutional neural networks for efficient and robust hand pose
  estimation from single depth images.
\newblock In {\em CVPR}, 2017.

\bibitem{geiger2013vision}
A.~Geiger, P.~Lenz, C.~Stiller, and R.~Urtasun.
\newblock Vision meets robotics: The kitti dataset.
\newblock {\em IJRR}, 32(11):1231--1237, 2013.

\bibitem{guo20153d}
K.~Guo, D.~Zou, and X.~Chen.
\newblock 3d mesh labeling via deep convolutional neural networks.
\newblock {\em TOG}, 35(1):3, 2015.

\bibitem{gupta2014learning}
S.~Gupta, R.~Girshick, P.~Arbel{\'a}ez, and J.~Malik.
\newblock Learning rich features from rgb-d images for object detection and
  segmentation.
\newblock In {\em ECCV}, 2014.

\bibitem{gupta2016cross}
S.~Gupta, J.~Hoffman, and J.~Malik.
\newblock Cross modal distillation for supervision transfer.
\newblock In {\em CVPR}, 2016.

\bibitem{hao2017scale}
Z.~Hao, Y.~Liu, H.~Qin, J.~Yan, X.~Li, and X.~Hu.
\newblock Scale-aware face detection.
\newblock In {\em CVPR}, 2017.

\bibitem{he2016deep}
K.~He, X.~Zhang, S.~Ren, and J.~Sun.
\newblock Deep residual learning for image recognition.
\newblock In {\em CVPR}, 2016.

\bibitem{hoiem2011representations}
D.~Hoiem and S.~Savarese.
\newblock Representations and techniques for 3d object recognition and scene
  interpretation.
\newblock {\em Synthesis Lectures on Artificial Intelligence and Machine
  Learning}, 5(5):1--169, 2011.

\bibitem{hu2016finding}
P.~Hu and D.~Ramanan.
\newblock Finding tiny faces.
\newblock In {\em CVPR}, 2017.

\bibitem{huang2015analysis}
H.~Huang, E.~Kalogerakis, and B.~Marlin.
\newblock Analysis and synthesis of 3d shape families via deep-learned
  generative models of surfaces.
\newblock In {\em Computer Graphics Forum}, volume~34, pages 25--38, 2015.

\bibitem{huang2016point}
J.~Huang and S.~You.
\newblock Point cloud labeling using 3d convolutional neural network.
\newblock In {\em ICPR}, 2016.

\bibitem{johnson1999using}
A.~E. Johnson and M.~Hebert.
\newblock Using spin images for efficient object recognition in cluttered 3d
  scenes.
\newblock {\em TPAMI}, 21(5):433--449, 1999.

\bibitem{kalogerakis20163d}
E.~Kalogerakis, M.~Averkiou, S.~Maji, and S.~Chaudhuri.
\newblock 3d shape segmentation with projective convolutional networks.
\newblock In {\em CVPR}, 2017.

\bibitem{kamnitsas2017efficient}
K.~Kamnitsas, C.~Ledig, V.~F. Newcombe, J.~P. Simpson, A.~D. Kane, D.~K. Menon,
  D.~Rueckert, and B.~Glocker.
\newblock Efficient multi-scale 3d cnn with fully connected crf for accurate
  brain lesion segmentation.
\newblock {\em Medical image analysis}, 36:61--78, 2017.

\bibitem{krizhevsky2012imagenet}
A.~Krizhevsky, I.~Sutskever, and G.~E. Hinton.
\newblock Imagenet classification with deep convolutional neural networks.
\newblock In {\em NIPS}, 2012.

\bibitem{li2016vehicle}
B.~Li, T.~Zhang, and T.~Xia.
\newblock Vehicle detection from 3d lidar using fully convolutional network.
\newblock {\em arXiv:1608.07916}, 2016.

\bibitem{li2016lstm}
Z.~Li, Y.~Gan, X.~Liang, Y.~Yu, H.~Cheng, and L.~Lin.
\newblock Lstm-cf: Unifying context modeling and fusion with lstms for rgb-d
  scene labeling.
\newblock In {\em ECCV}, 2016.

\bibitem{lin2016efficient}
G.~Lin, C.~Shen, A.~van~den Hengel, and I.~Reid.
\newblock Efficient piecewise training of deep structured models for semantic
  segmentation.
\newblock In {\em CVPR}, 2016.

\bibitem{lin2016feature}
T.-Y. Lin, P.~Doll{\'a}r, R.~Girshick, K.~He, B.~Hariharan, and S.~Belongie.
\newblock Feature pyramid networks for object detection.
\newblock In {\em CVPR}, 2017.

\bibitem{long2015fully}
J.~Long, E.~Shelhamer, and T.~Darrell.
\newblock Fully convolutional networks for semantic segmentation.
\newblock In {\em CVPR}, 2015.

\bibitem{masci2015geodesic}
J.~Masci, D.~Boscaini, M.~Bronstein, and P.~Vandergheynst.
\newblock Geodesic convolutional neural networks on riemannian manifolds.
\newblock In {\em ICCV Workshops}, 2015.

\bibitem{maturana20153d}
D.~Maturana and S.~Scherer.
\newblock 3d convolutional neural networks for landing zone detection from
  lidar.
\newblock In {\em ICRA}, 2015.

\bibitem{orts2016holoportation}
S.~Orts-Escolano, C.~Rhemann, S.~Fanello, W.~Chang, A.~Kowdle, Y.~Degtyarev,
  D.~Kim, P.~L. Davidson, S.~Khamis, M.~Dou, et~al.
\newblock Holoportation: Virtual 3d teleportation in real-time.
\newblock In {\em UIST}, 2016.

\bibitem{qi2016pointnet}
C.~R. Qi, H.~Su, K.~Mo, and L.~J. Guibas.
\newblock Pointnet: Deep learning on point sets for 3d classification and
  segmentation.
\newblock In {\em CVPR}, 2017.

\bibitem{qi2016volumetric}
C.~R. Qi, H.~Su, M.~Nie{\ss}ner, A.~Dai, M.~Yan, and L.~J. Guibas.
\newblock Volumetric and multi-view cnns for object classification on 3d data.
\newblock In {\em CVPR}, 2016.

\bibitem{qi2017pointnetpp}
C.~R. Qi, L.~Yi, H.~Su, and L.~J. Guibas.
\newblock Pointnet++: Deep hierarchical feature learning on point sets in a
  metric space.
\newblock In {\em NIPS}, 2017.

\bibitem{qi20173d}
X.~Qi, R.~Liao, J.~Jia, S.~Fidler, and R.~Urtasun.
\newblock 3d graph neural networks for rgbd semantic segmentation.
\newblock In {\em CVPR}, 2017.

\bibitem{riegler2016octnet}
G.~Riegler, A.~O. Ulusoys, and A.~Geiger.
\newblock Octnet: Learning deep 3d representations at high resolutions.
\newblock {\em arXiv:1611.05009}, 2016.

\bibitem{sedaghat2016orientation}
N.~Sedaghat, M.~Zolfaghari, and T.~Brox.
\newblock Orientation-boosted voxel nets for 3d object recognition.
\newblock {\em arXiv:1604.03351}, 2016.

\bibitem{shi2015deeppano}
B.~Shi, S.~Bai, Z.~Zhou, and X.~Bai.
\newblock Deeppano: Deep panoramic representation for 3-d shape recognition.
\newblock {\em Signal Processing Letters}, 22(12):2339--2343, 2015.

\bibitem{silberman2012indoor}
N.~Silberman, D.~Hoiem, P.~Kohli, and R.~Fergus.
\newblock Indoor segmentation and support inference from rgbd images.
\newblock In {\em ECCV}, 2012.

\bibitem{simonyan2014very}
K.~Simonyan and A.~Zisserman.
\newblock Very deep convolutional networks for large-scale image recognition.
\newblock {\em arXiv:1409.1556}, 2014.

\bibitem{song2016deep}
S.~Song and J.~Xiao.
\newblock Deep sliding shapes for amodal 3d object detection in rgb-d images.
\newblock In {\em CVPR}, 2016.

\bibitem{song2016semantic}
S.~Song, F.~Yu, A.~Zeng, A.~X. Chang, M.~Savva, and T.~Funkhouser.
\newblock Semantic scene completion from a single depth image.
\newblock In {\em CVPR}, 2017.

\bibitem{su2015multi}
H.~Su, S.~Maji, E.~Kalogerakis, and E.~Learned-Miller.
\newblock Multi-view convolutional neural networks for 3d shape recognition.
\newblock In {\em ICCV}, 2015.

\bibitem{tchapmi2017segcloud}
L.~P. Tchapmi, C.~B. Choy, I.~Armeni, J.~Gwak, and S.~Savarese.
\newblock Segcloud: Semantic segmentation of 3d point clouds.
\newblock {\em arXiv:1710.07563}, 2017.

\bibitem{uhrig2017sparsity}
J.~Uhrig, N.~Schneider, L.~Schneider, U.~Franke, T.~Brox, and A.~Geiger.
\newblock Sparsity invariant cnns.
\newblock In {\em 3DV}, 2017.

\bibitem{wang2016torontocity}
S.~Wang, M.~Bai, G.~Mattyus, H.~Chu, W.~Luo, B.~Yang, J.~Liang, J.~Cheverie,
  S.~Fidler, and R.~Urtasun.
\newblock Torontocity: Seeing the world with a million eyes.
\newblock In {\em ICCV}, 2017.

\bibitem{wang2016autoscaler}
S.~Wang, L.~Luo, N.~Zhang, and J.~Li.
\newblock Autoscaler: Scale-attention networks for visual correspondence.
\newblock In {\em BMVC}, 2017.

\bibitem{wu20153d}
Z.~Wu, S.~Song, A.~Khosla, F.~Yu, L.~Zhang, X.~Tang, and J.~Xiao.
\newblock 3d shapenets: A deep representation for volumetric shapes.
\newblock In {\em CVPR}, 2015.

\bibitem{xie2015deepshape}
J.~Xie, Y.~Fang, F.~Zhu, and E.~Wong.
\newblock Deepshape: Deep learned shape descriptor for 3d shape matching and
  retrieval.
\newblock In {\em CVPR}, 2015.

\bibitem{xu2016multimodal}
P.~Xu, F.~Davoine, J.-B. Bordes, H.~Zhao, and T.~Den{\oe}ux.
\newblock Multimodal information fusion for urban scene understanding.
\newblock {\em Machine Vision and Applications}, 27(3):331--349, 2016.

\bibitem{zaheer2017deep}
M.~Zaheer, S.~Kottur, S.~Ravanbakhsh, B.~Poczos, R.~Salakhutdinov, and
  A.~Smola.
\newblock Deep sets.
\newblock In {\em NIPS}, 2017.

\bibitem{zeng20163dmatch}
A.~Zeng, S.~Song, M.~Nie{\ss}ner, M.~Fisher, and J.~Xiao.
\newblock 3dmatch: Learning the matching of local 3d geometry in range scans.
\newblock In {\em CVPR}, 2017.

\bibitem{zhang2017scale}
R.~Zhang, S.~Tang, Y.~Zhang, J.~Li, and S.~Yan.
\newblock Scale-adaptive convolutions for scene parsing.
\newblock In {\em ICCV}, 2017.

\bibitem{zhang2016deepcontext}
Y.~Zhang, M.~Bai, P.~Kohli, S.~Izadi, and J.~Xiao.
\newblock Deepcontext: context-encoding neural pathways for 3d holistic scene
  understanding.
\newblock In {\em ICCV}, 2017.

\bibitem{zhao2016pyramid}
H.~Zhao, J.~Shi, X.~Qi, X.~Wang, and J.~Jia.
\newblock Pyramid scene parsing network.
\newblock In {\em CVPR}, 2017.

\bibitem{zhou2014object}
B.~Zhou, A.~Khosla, A.~Lapedriza, A.~Oliva, and A.~Torralba.
\newblock Object detectors emerge in deep scene cnns.
\newblock {\em arXiv:1412.6856}, 2014.

\end{thebibliography}
}
	
\end{document}